\documentclass[letterpaper, 10 pt, conference]{ieeeconf}  

\IEEEoverridecommandlockouts
\usepackage{graphicx}
\usepackage{amsmath}
\usepackage{amssymb}
\usepackage{cite}
\usepackage{url}
\usepackage{microtype}

\title{\LARGE \bf
Robust 2D Traversability Mapping for Construction AMRs via Failure-Mode-Aware Fusion of LiDAR Geometry and Monocular Semantics
}

\author{Manoj Karnekar$^{1}$, Om Mandhane$^{2}$, and Gautham Ramkumar$^{3}$
\thanks{$^{1}$Manoj Karnekar is Robotics Lead at FloMobility Pvt. Ltd., Bangalore, India
        {\tt\small manojkarnekar1@gmail.com}}%
\thanks{$^{2}$Om Mandhane is a Robotics Engineer at FloMobility Pvt. Ltd., Bangalore, India
        {\tt\small ommandhane27@gmail.com}}%
\thanks{$^{3}$Gautham Ramkumar is a Robotics Engineer at FloMobility Pvt. Ltd., Bangalore, India
        {\tt\small gauthamramkumar03@gmail.com}}%
}

\begin{document}

\maketitle
\thispagestyle{empty}
\pagestyle{empty}

\begin{abstract}

Autonomous Mobile Robots (AMRs) on active construction sites face severe navigational challenges: geometry-based traversability mapping (e.g., LiDAR) misses visually hazardous but geometrically flat surfaces like wet mud and ponding concrete, while abrupt geometry on drivable speed-breakers and inclines produces phantom obstacles. We propose a real-time, failure-mode-aware multimodal traversability pipeline on an NVIDIA Jetson AGX Orin, where LiDAR is the primary geometric safety estimate and monocular semantics act as a selective, class- and confidence-gated corrective signal. The representation retains distinct traversable classes, namely flat road, terrain, and rocky terrain, while flagging construction hazards. We also release a multimodal construction-site dataset from a custom AMR: four closed-loop ROS~2 sequences from two active sites (RGB, depth, LiDAR, IMU, GPS-RTK, odometry) plus 506 annotated frames across 28 semantic classes. By projecting LiDAR onto dense semantic masks, resolving sparsity via morphological in-painting, and applying failure-mode-aware fusion with Patchwork++, the system corrects complementary geometric failure modes for a local AMR costmap.

\end{abstract}

\section{INTRODUCTION}

AMR deployment on active construction sites is hindered by unstructured terrain: amorphous hazards (deep mud, ponding concrete) and transient structures (scaffolding, rebar, temporary planks) violate structured-indoor-navigation assumptions. AMRs conventionally rely on 3D LiDAR for ground segmentation (e.g., Patchwork++ \cite{lee2022patchwork}), which suffers two complementary failure modes here: (1) false negatives, where flat non-load-bearing hazards like wet mud or ponding concrete read as safe, traversable ground ($Z \approx 0$); and (2) false positives, where ground segmentation fails right after drivable elevation changes such as cresting a speed-breaker or climbing a steep earthen incline: the sharp negative gradient at the trailing edge is misread as a lethal obstacle (a ``black line'' in the costmap), causing navigational freezing. Purely semantic (camera-based) approaches address recognition but lack the metric reliability collision avoidance needs \cite{schilling2017,zhou2022,liu2023bevfusion,zhu2024learning,frey2024roadrunner}.

This motivates a fusion strategy in which the two sensors are not treated as interchangeable sources, but are assigned complementary roles according to their known failure modes. Prior fusion and learned approaches (Section~\ref{sec:related}) show the benefit of combining semantic appearance with geometric structure, but leave open the application-specific problem of assigning traversability to the resulting terrain representation.

We propose a failure-mode-aware semantic-geometric fusion architecture where LiDAR remains the primary safety estimate and a YOLOv26-m-seg model \cite{ultralytics2026} provides class- and confidence-gated corrective evidence, overriding geometry only for identified failure modes. Rather than learn the final traversability decision end-to-end \cite{zhu2024learning,seo2023metaverse}, we use learning only for semantic recognition and keep sensor arbitration an explicit, auditable, deterministic policy. This suits our 506-image annotated set, which supervises recognition directly (a learned traversability predictor would need an additional target describing the desired traversability of robot-centric terrain regions, which existing learned approaches obtain from proprioceptive or self-supervised signals \cite{zhu2024learning,seo2023metaverse}), and it gives the per-decision verifiability a safety-relevant system needs. The final costmap is derived deterministically from semantic evidence and measured LiDAR geometry, so the fusion logic can be designed explicitly around the construction-site failure modes considered in this work, and the result feeds our custom path-planning and navigation stack.

\textbf{Contributions:}
\begin{enumerate}
\item A deterministic, class- and confidence-gated sensor-arbitration strategy using semantic vision as a selective corrective signal for complementary LiDAR failure modes (flat hazards, phantom obstacles), with LiDAR as fallback. It addresses both geometric false negatives (flat wet mud, ponding concrete) and geometric false positives (speed-breakers, inclined terrain).
\item The public multimodal construction-site dataset detailed in Section~\ref{sec:dataset}: four closed-loop ROS~2 sequences from two active sites (RGB, depth, LiDAR, IMU, GPS-RTK, odometry) plus 506 pixel-level annotated RGB frames spanning 28 construction-specific semantic classes. The taxonomy distinguishes traversable surfaces (flat road, terrain, rocky terrain) from construction-specific hazards, while the raw recordings preserve the temporal structure of robotic operation.
\end{enumerate}
We validate the framework via cross-site evaluation, training on one site and testing perception and fusion jointly on an unseen site, with particular emphasis on the complementary geometric failure modes addressed by the proposed fusion strategy.

\section{RELATED WORK}
\label{sec:related}

\subsection{Geometric Ground Segmentation}

Geometric approaches such as Patchwork++ \cite{lee2022patchwork} give robust, real-time ground-plane extraction from raw 3D LiDAR by adaptively fitting region-wise planes and rejecting partial under-segmentation; they are computationally efficient and metrically reliable, and our geometric branch is built directly on this approach. However, because they reason purely over spatial gradients and elevation, they cannot distinguish flat non-load-bearing hazards (e.g., wet mud) from safe ground, nor drivable elevation changes (e.g., ramps, speed-breakers) from lethal obstacles---the two central failure modes this paper addresses.

\subsection{Off-Road and Urban Semantic Datasets}

Large-scale urban datasets such as Cityscapes \cite{cordts2016cityscapes} established the benchmark methodology for pixel-level semantic segmentation, but their class taxonomy and structured-road assumptions do not transfer to unstructured terrain. Off-road semantic segmentation datasets such as RUGD \cite{wigness2019rugd} and RELLIS-3D \cite{jiang2021rellis} extend the Cityscapes paradigm to unstructured environments, introducing fine-grained terrain classes collected from mobile robots traversing natural environments; RELLIS-3D additionally provides multimodal LiDAR, camera, and inertial data in a ROS-based dataset. Neither, however, captures the transient, man-made hazard classes specific to active construction sites, such as rebar, scaffolding, ponding concrete, or loose pipe stacks.

TNS's Complex Worksite Terrain (CWT) dataset \cite{guan2022tns} targets excavators with 669 RGB-only images and seven categories based on terrain and navigability. Our release instead focuses on live, active building construction sites: rather than primarily an annotated RGB corpus, it centers on a raw multimodal rosbag corpus (four closed-loop sequences: RGB, depth, LiDAR, IMU, GPS-RTK, odometry) plus a 28-class annotated subset, supporting fusion, mapping, loop-closure, and localization research along with segmentation. The taxonomy explicitly separates construction hazards and traversable surface types that are operationally important for a compact AMR.

\subsection{Semantic-Geometric Fusion}

Prior fusion work combines visual and geometric terrain cues \cite{schilling2017,zhou2022}: Schilling et al.\ combine geometric and visual terrain information for autonomous mobile navigation, while Zhou et al.\ investigate terrain traversability mapping using LiDAR and camera fusion. These works establish the benefit of combining visual appearance with geometric terrain information. More recent learned approaches (BEVFusion \cite{liu2023bevfusion}, RoadRunner \cite{frey2024roadrunner}, METAVerse \cite{seo2023metaverse}) learn multimodal or traversability representations end-to-end: BEVFusion provides a unified bird's-eye-view representation, while RoadRunner and METAVerse demonstrate learned traversability estimation from multimodal observations or vehicle-terrain interaction. In contrast, our approach uses learning only for semantic recognition and keeps the final sensor arbitration a deterministic policy, so the fusion rules can be tied directly to the construction-specific LiDAR failure modes observed in our data.

\subsection{Geometric-Semantic Fusion for Construction Machinery}

TNS \cite{guan2022tns} is closest to our work, fusing elevation-grid slope/step-height features with RGB semantics to zero traversability for non-traversable categories on a 49-ton excavator. TNS takes a 3D LiDAR point cloud, an RGB camera stream, and the robot pose, constructs an elevation grid, estimates terrain features such as slope and step height, and incorporates semantic information into the resulting traversability representation. Its semantic-geometric fusion is explicit and deterministic: non-traversable semantic categories can be assigned zero traversability, while other regions retain or receive a geometric traversability score.

Deterministic semantic-geometric fusion itself is therefore not claimed as the novelty of our work, and the distinction from TNS is not that our method is deterministic while TNS is learned. The distinction lies in how semantic information is used in the final decision. TNS primarily produces a continuous traversability representation based on geometric terrain properties and semantic categories for excavator planning. Where TNS applies a uniform semantic override, our fusion logic is asymmetric, failure-mode-specific, and confidence-gated, treating semantic perception as a corrective signal for a local AMR costmap, tailored to the unstructured hazards of active building sites: high-confidence flat hazards override geometrically clear LiDAR cells, high-confidence traversable classes suppress geometric false positives from speed-breakers and ramps, and ambiguous classes (e.g., puddles) defer to LiDAR. For puddles, semantic perception confirms water presence but does not independently determine traversability; LiDAR geometry remains authoritative. We also preserve flat road, terrain, and rocky terrain as distinct states rather than a single traversable class, enabling surface-specific navigation costs and speed preferences.

\section{CONSTRUCTION-SITE MULTIMODAL DATASET}
\label{sec:dataset}
\begin{figure}[t]
   \centering
   \includegraphics[width=0.75\columnwidth]{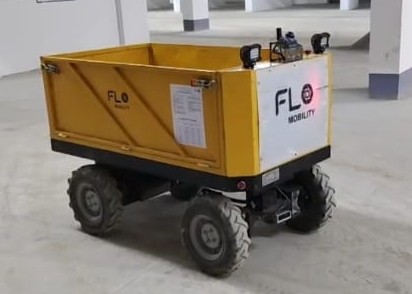}
   \vspace{-0.3em}
   \caption{The custom AMR platform used for multimodal data collection, purpose-built for autonomous navigation in construction environments.}
   \label{fig_robot}
   \vspace{-0.5em}
\end{figure}

To train the semantic branch of our perception stack and support multimodal robotics research, we curated a dataset explicitly for unstructured construction environments. The release contains both the underlying multimodal robotic recordings and a curated pixel-level annotated semantic subset; the 506 annotated images are a supervised subset of a substantially larger temporal sensor corpus.

\subsection{Data Collection and Keyframe Extraction}

The data was gathered at two active construction sites (Site~1, Site~2) using a custom AMR (Fig.~\ref{fig_robot}) equipped with an OAK-D PoE Pro RGB-D camera, a Livox Mid-360 LiDAR, IMU, GPS-RTK, and odometry, and stored in ROS~2 MCAP bags for a total of four closed-loop sessions (2 per site; $>$1.75~h, 6,300~s of continuous field operation), capturing more than 3.06M multimodal ROS messages. In addition to the 506 annotated frames used here, the raw corpus contains spatial regions that can be revisited and temporal relationships between streams that can be maintained; it can also be used for point-cloud generation, SLAM/loop-closure, GPS-RTK localization, sensor calibration studies, and more hazard-class coverage.

To build a high-variance training set and avoid the redundancy of standard temporal extraction, a perceptual-hash (pHash) keyframe extractor is used to extract key frames by selecting those with high variance in RGB frames. Each candidate RGB frame is passed through the pHash algorithm to generate a low-frequency structural fingerprint, and the Hamming distance $D_H$ between the candidate's hash $h_c$ and the hashes $h_s$ of all previously saved frames is computed. A candidate is retained iff
\begin{equation}
\min\big(D_H(h_c,h_s)\big)\geq \tau,
\end{equation}
where $\tau{=}14$ is an empirically chosen distance threshold, so that every retained keyframe differs from all previously saved ones by at least $\tau$ in hash space.

Of the more than 3.06M raw multimodal messages captured across all four closed-loop sessions, only 506 keyframes have been pixel-level annotated and used to train the semantic branch reported in this paper. The remaining rosbag corpus is part of the public dataset and can support future dataset scaling, active-learning-based annotation, and temporal-consistency studies.

\subsection{Dataset Release}

The release contains the four raw ROS~2 MCAP recordings with their metadata and calibration information, and the 506 annotated RGB frames with their pixel-level semantic masks. The fine-tuned YOLOv26-m-seg checkpoint, training configuration, and class taxonomy are also released to enable reproduction of the semantic segmentation results reported here. The raw recordings preserve RGB, depth, LiDAR, IMU, GPS-RTK, odometry, and timestamp information, enabling tasks beyond semantic segmentation such as multimodal sensor fusion, semantic point-cloud generation, and navigation research.

The annotated dataset, calibration data, and trained model are available at \url{https://huggingface.co/datasets/manojkarnekar/construction-traversability-dataset} \cite{karnekar2026dataset}, while the raw ROS~2 MCAP recordings are released separately at \url{https://huggingface.co/datasets/manojkarnekar/construction-rosbags} \cite{karnekar2026rosbags}. To address privacy concerns, people, text, and vehicle license plates are blurred before release.

\subsection{Class Taxonomy and Fusion Roles}

Upon analyzing the real-world construction-site data, we defined a 28-class semantic taxonomy covering both general scene elements and construction-specific terrain and hazard categories. Rather than treating all classes identically during fusion, the taxonomy assigns each class a fusion role according to its expected traversability and the failure mode it addresses (Table~\ref{table_taxonomy}): three \emph{traversable} classes (flat road, terrain, rocky terrain); \emph{flat-hazard override} classes (wet mud, ponding concrete) that can override geometrically clear LiDAR cells; a \emph{geometry-dependent} class (puddle) confirming water but deferring depth to LiDAR; \emph{camera-ignored} classes (sky, not visible) that always fall back to LiDAR; and 20 remaining \emph{non-traversable} classes (e.g., rebar, scaffolding, debris, pipes, vehicles, personnel).

The three traversable classes are kept as distinct semantic states rather than collapsed into a binary representation, preserving surface-specific information for downstream cost and speed assignment. Wet mud and ponding concrete are visually identifiable hazards that can remain geometrically flat and therefore constitute important LiDAR false-negative cases; they may override a geometrically clear cell only when their semantic confidence exceeds a dedicated hazard threshold. Puddle is treated differently: semantic perception confirms the presence of water, but LiDAR geometry remains responsible for determining whether the region is physically traversable. Sky and not visible indicate regions where camera semantics should not independently determine traversability, so they fall back directly to the LiDAR estimate. All remaining classes represent visually identified non-traversable objects or surfaces and participate in the confidence-gated semantic correction of geometrically clear regions.

\begin{table}[t]
\caption{28-Class Construction-Site Semantic Taxonomy and Fusion Roles}
\label{table_taxonomy}
\vspace{-0.5em}
\begin{center}
\small
\begin{tabular}{|p{2.0cm}|p{5.3cm}|}
\hline
\textbf{Fusion role} & \textbf{Semantic classes} \\
\hline
Traversable & flat road, terrain, rocky terrain \\
\hline
Flat-hazard override & wet mud, ponding concrete \\
\hline
Geometry-dep. hazard & puddle \\
\hline
Camera-ignored & not visible, sky \\
\hline
Non-traversable (20 classes) & animal, building, ceiling, concrete blocks, construction machinery, debris, fence, material pile, person, pillar, pipes, pit, pole, rebar, scaffolding, tiles, vegetation, vehicle, wall, wooden planks \\
\hline
\end{tabular}
\end{center}
\vspace{-1em}
\end{table}

\begin{figure*}[t]
   \centering
   \includegraphics[width=0.85\textwidth]{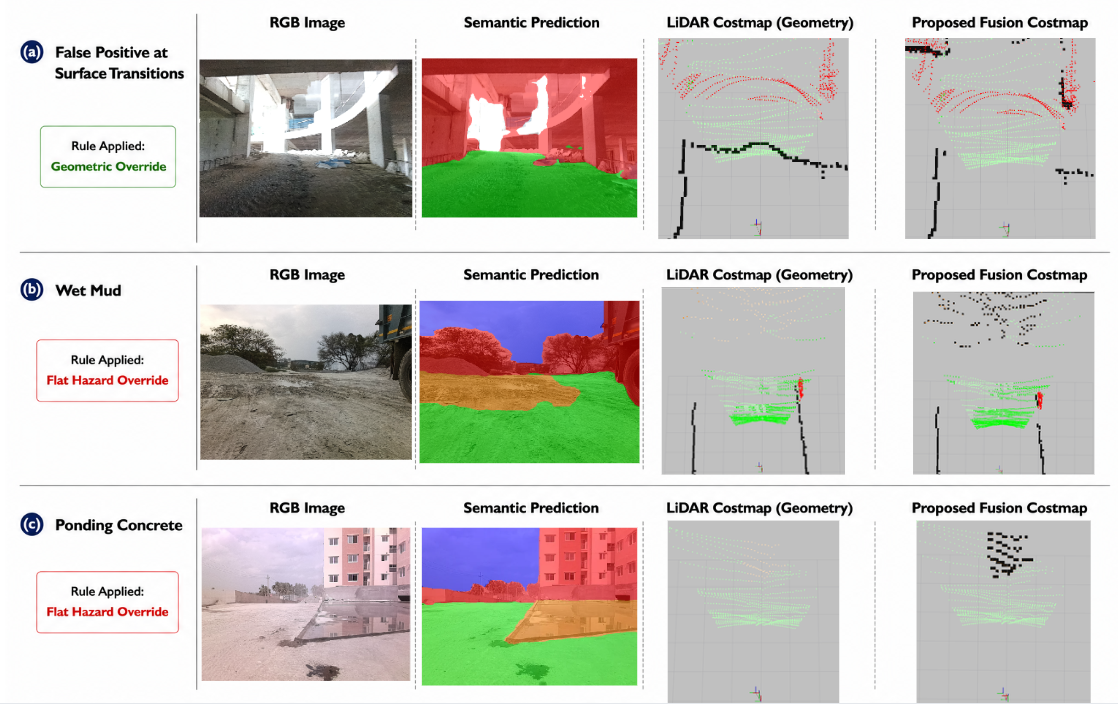}
   \vspace{-0.3em}
   \caption{Qualitative fusion results. (a) LiDAR falsely flags a traversable surface transition, such as a slope or an incline, as an obstacle; the flat road prediction removes it (geometric override). (b--c) Wet mud and ponding concrete are geometrically flat but semantically identified, triggering the flat-hazard override.}
   \label{fig_qual}
   \vspace{-0.5em}
\end{figure*}

\subsection{Automated Hybrid Annotation Pipeline}

An automated hybrid annotation pipeline followed by human validation was used to add 506 keyframes to the $>$3.06M messages (415 train / 91 val):
\begin{enumerate}
\item \textbf{Hazard mask proposal (SAM-3):} SAM-3 \cite{carion2025sam3}, built on the promptable segmentation paradigm introduced by the original Segment Anything Model \cite{kirillov2023sam}, automatically isolates and masks distinct physical hazard classes.
\item \textbf{Background classes (YOLO):} for broad environmental classes, a YOLO-based model \cite{ultralytics2026} maps background contexts, which are programmatically merged with the SAM-3 masks.
\item \textbf{Human-in-the-loop validation:} human annotators perform final validation and correction of the generated masks to ensure class fidelity.
\end{enumerate}

\subsection{Spatial Generalization Split}

While raw telemetry was gathered from both locations, the annotated training data is only from Site~1, and Site~2 is reserved for evaluation on the zero-shot spatial-generalization task, i.e., as a deliberately unseen environment. The closed-loop nature of all four recordings is retained in the public release, although the site-level split is the primary protocol used for the semantic and fusion evaluation reported here.

\section{SYSTEM ARCHITECTURE AND METHODOLOGY}

The system runs a dual-stream perception pipeline into a unified 2D occupancy grid for the AMR's navigation stack \cite{lu2014costmap}.

\subsection{Semantic Branch (YOLOv26-m-seg)}

For real-time edge deployment on the NVIDIA Jetson AGX Orin, we selected the YOLOv26-m-seg model \cite{ultralytics2026}. We take the semantic head's logits, bilinearly upsample them to full image resolution, and apply a softmax over the 28 classes, so every pixel carries a class label together with its posterior probability as a confidence value; the model thus outputs a dense per-pixel class map with corresponding confidence scores.

\subsection{Geometric Branch (Patchwork++)}

The \emph{geometric branch} (Patchwork++ \cite{lee2022patchwork}, tuned for the Livox Mid-360's $23^{\circ}$ mounting tilt) segments the cloud into ground and non-ground returns, estimating ground planes and flagging geometric obstacles from spatial gradients and step heights; non-ground returns are rasterized into the grid as geometric obstacles. RGB and LiDAR messages are paired by header timestamp with a 0.1~s tolerance.

\subsection{Semantic Point Cloud Generation via Intrinsic Projection}

To fuse the modalities, the 3D geometric points must be colored with 2D semantic classes. A LiDAR point $P_{bf}=[X_{bf},Y_{bf},Z_{bf}]^{\top}$ in the robot base frame is transformed to the camera's optical frame via the extrinsic calibration $T^{lidar}_{cam}$, giving $P_c=[X,Y,Z]^{\top}$, and projected using the pinhole model with the sensor's intrinsic matrix,
\begin{equation}
u = \left\lfloor f_x \tfrac{X}{Z} + c_x \right\rfloor,\ v = \left\lfloor f_y \tfrac{Y}{Z} + c_y \right\rfloor,
\end{equation}
where $(f_x,f_y)$ are the focal lengths and $(c_x,c_y)$ the optical center. The point inherits the semantic label and confidence at pixel $(u,v)$ when $Z\!>\!0$ and $(u,v)$ is in-bounds ($0\le u<W$, $0\le v<H$ for an image of width $W$ and height $H$).

\subsection{Grid Flattening and Morphological In-Painting}

The labeled points are flattened into a $200\!\times\!200$, 0.1~m local grid ($20\!\times\!20$~m, robot-centered in \texttt{base\_footprint}) initialized to Unknown ($-1$). Returns above the robot height (1.30~m) are discarded, and where several points share a cell the highest-confidence label wins. Since sparse LiDAR rings leave gaps, we resolve this without computationally heavy neural upsampling: for each Unknown cell we count confident same-tier hits in a $9\!\times\!9$ neighborhood and fill the cell once support is sufficient, requiring one hit for flat road and two for terrain/rocky terrain, at a confidence just above the corresponding override threshold. The stricter tier-2 rule keeps in-painting from propagating the weaker-evidence surface classes.

\subsection{Failure-Mode-Aware Deterministic Fusion Engine}
\label{sec:fusion}

The semantic classes in Table~\ref{table_taxonomy} are not treated uniformly during fusion. For every cell of the 2D local costmap, the system evaluates the LiDAR geometric state, the semantic class label, and the network confidence, and applies class-specific arbitration rules designed around the expected failure mode of each class group. Fusion applies this class- and confidence-gated arbitration as a strict, deterministic, auditable decision tree per cell:
\begin{enumerate}
\item \textbf{Sensor blindness:} Sky/Not Visible $\rightarrow$ the camera is ignored; LiDAR only.
\item \textbf{Flat-hazard override:} Wet Mud/Ponding Concrete with conf~$>$~0.60 $\rightarrow$ force Obstacle, overriding flat LiDAR readings; below this threshold the cell defers to LiDAR.
\item \textbf{Depth verification:} Puddle $\rightarrow$ the camera confirms water, and the depth/traversability decision is deferred to LiDAR.
\item \textbf{Sensor agreement:} if LiDAR and camera agree $\rightarrow$ output as-is.
\item \textbf{Correction of geometric false negatives:} LiDAR Clear + camera non-traversable with conf~$>$~0.80 $\rightarrow$ camera overrides; otherwise geometry is trusted.
\item \textbf{Correction of geometric false positives (speed-breakers/ramps):} LiDAR Obstacle + camera Flat Road (conf~$>$~0.40) or Terrain/Rocky Terrain (conf~$>$~0.50) $\rightarrow$ output Clear.
\item \textbf{Traversable surface differentiation:} the label grid keeps the three traversable classes distinct, allowing the downstream navigation stack to assign different traversal costs or speed preferences according to surface type.
\item \textbf{Fallback:} all other cases fall back to LiDAR geometric safety.
\end{enumerate}

\section{EXPERIMENTS AND RESULTS}

\subsection{Semantic Model Training Performance}

The YOLOv26-m-seg model, fine-tuned on 415/91 train/val images, achieved an mIoU of 0.579 and a pixel accuracy of 82.4\%. Structurally dominant classes performed well (Sky 0.953, Vegetation 0.858, Flat Road 0.816 mIoU), while rare hazards suffered from sparsity (Wet Mud 0.188 mIoU). Low mIoU on these large, amorphous regions is dominated by boundary error, whereas the flat-hazard override only needs confident interior cells; this is why Wet Mud still triggers reliably downstream (7/7 cases across Tables~\ref{table_fmea_sri} and~\ref{table_fmea_sow}) despite its weak segmentation score.

\subsection{Runtime and Edge Deployment Benchmark}

To quantify real-time feasibility on the target hardware, we ran two complementary benchmarks on the Jetson AGX Orin: (i) standalone semantic segmentation inference and (ii) the integrated RGB--LiDAR semantic fusion and costmap pipeline (Table~\ref{table_runtime}).

\textbf{Standalone inference.} We benchmarked standalone YOLOv26-m-seg inference at $640\!\times\!640$ on the CUDA GPU, using the 415 readable RGB images of the training subset for timing purposes only (no accuracy metric is computed on this set), with 50 warm-up iterations followed by 500 measured iterations. This yields a mean latency of 23.18~ms (median 23.04~ms, P95 23.93~ms, P99 25.49~ms), a mean throughput of 43.14~FPS, well within the budget of the downstream 10~Hz costmap update.

\begin{table}[t]
\caption{Runtime Benchmark on the NVIDIA Jetson AGX Orin (ms)}
\label{table_runtime}
\vspace{-0.5em}
\begin{center}
\footnotesize
\begin{tabular}{|l|c|c|c|c|}
\hline
\textbf{Benchmark} & \textbf{Mean} & \textbf{Median} & \textbf{P95} & \textbf{P99} \\
\hline
YOLOv26-m-seg & 23.18 & 23.04 & 23.93 & 25.49 \\
Integrated computation & 88.96 & 88.95 & 111.60 & 118.95 \\
\hline
\end{tabular}
\end{center}
\vspace{-1em}
\end{table}

\textbf{Integrated pipeline.} The integrated pipeline was benchmarked while processing ROS~2 sensor data from the recorded construction-site sequences. For each valid sample the computational latency is
\begin{equation}
T_{comp} = T_{YOLO} + T_{proj} + T_{fusion},
\end{equation}
covering YOLO semantic inference, LiDAR projection and semantic point-cloud generation, and deterministic costmap fusion. ROS synchronization wait time is excluded, since it reflects inter-node scheduling and synchronization delay rather than computation performed by the perception and fusion modules. Warm-up samples and samples for which the fusion node reported stale semantic or geometric data were also excluded, leaving 84 valid samples. Across them, the mean end-to-end costmap latency was 88.96~ms (median 88.95~ms, P95 111.60~ms, P99 118.95~ms), corresponding to a mean throughput of 11.24~Hz. The costmap is configured to update at 10~Hz (a 100~ms period), so the mean is below the period but the tail does exceed the 100~ms cycle time: the pipeline has sufficient mean computational capacity for the intended rate, while occasional longer-latency updates remain observable in the tail. This does not stall the planner: the fusion node publishes on a fixed 10~Hz timer decoupled from the perception callbacks, so a late cycle republishes with a marginally staler semantic layer rather than a missed costmap, and a 0.3~s staleness guard flags the case where the geometric and semantic grids drift further apart than that.

The standalone and integrated benchmarks measure different aspects of the deployed system: the former isolates the inference cost of the fine-tuned YOLOv26-m-seg model, whereas the latter measures the combined computational cost of semantic inference, LiDAR projection and semantic point-cloud generation, and deterministic costmap fusion within the ROS~2 perception pipeline.

\subsection{Qualitative Traversability Evaluation}

Fig.~\ref{fig_qual} presents representative examples of the complementary failure modes addressed by the fusion strategy, and illustrates both corrective mechanisms: geometric override removes phantom obstacles at traversable transitions, and flat-hazard override assigns non-traversable cost to visually identifiable but geometrically flat hazards. At surface transitions, such as inclined slopes or the trailing edge of a speed-breaker, the LiDAR-based geometric costmap can produce a spurious obstacle although the underlying surface is traversable; there, the semantic prediction identifies the region as flat road, allowing the geometric override to remove the false obstacle from the fused costmap. Conversely, flat hazards such as wet mud and ponding concrete can appear traversable to LiDAR because they exhibit limited geometric variation, and the semantic branch supplies the missing appearance-based information. Together, these highlight the asymmetric nature of the fusion strategy: semantic evidence is used selectively to correct known geometric failure modes. At the same time, LiDAR remains the primary safety estimate for unresolved cases.

\subsection{Operational Failure Mode and Effects Analysis (FMEA)}
\label{sec:fmea}

The primary evaluation is cross-site: the semantic branch trains only on Site~1, and the full system is scored on unseen Site~2 (Table~\ref{table_fmea_sow}); Table~\ref{table_fmea_sri} additionally reports same-site Site~1 cases per failure mode. The scenarios in both tables were manually identified through review of the recorded rosbag footage, targeting instances of the specific LiDAR failure modes the fusion policy is designed to address (Section~\ref{sec:fusion}). This evaluation is intended as a failure-mode stress test rather than a statistically representative random sample; a broader quantitative evaluation of the fused costmap over randomly sampled frames is left to future work. Both tables confirm that the fusion engine resolves most targeted failure modes (inclined slopes, speedbreakers, wet mud, ponding concrete, debris, rebar, pipes) against the ground truth, with the camera column showing the raw semantic prediction that fusion arbitrates against LiDAR, while also surfacing a systematic failure mode discussed in Section~\ref{sec:limitations}, where a semantic classification error propagates through, rather than being suppressed by, the fusion policy.

\begin{table}[t]
\caption{FMEA Validation -- Site 1 (Primary/Training)}
\label{table_fmea_sri}
\vspace{-0.5em}
\begin{center}
\scriptsize
\begin{tabular}{|l|c|l|c|c|c|}
\hline
\textbf{Scenario} & \textbf{N} & \textbf{Camera} & \textbf{LiDAR} & \textbf{GT} & \textbf{Fused} \\
\hline
Inclined slope & 6 & Flat Road & NT & T & \textbf{T} \\
Speedbreaker & 8 & Flat Road & NT & T & \textbf{T} \\
Random LiDAR fail & 5 & Flat Road & NT & T & \textbf{T} \\
Debris & 7 & Debris & T & NT & \textbf{NT} \\
Metal pipe stack & 4 & Pipes & T & NT & \textbf{NT} \\
Concrete block & 3 & Debris & T & NT & \textbf{NT} \\
Weak wooden plank & 1 & Wooden Plank & T & NT & \textbf{NT} \\
Rebar & 3 & Rebar & T & NT & \textbf{NT} \\
Ponding concrete & 2 & Ponding Conc. & T & NT & \textbf{NT} \\
Bushes/vegetation & 3 & Vegetation & T & NT & \textbf{NT} \\
Animal & 1 & Animal & T & NT & \textbf{NT} \\
Wet mud & 3 & Wet Mud & T & NT & \textbf{NT} \\
\hline
\end{tabular}
\end{center}
\vspace{-0.6em}
\begin{center}
\footnotesize{T = Traversable, NT = Non-Traversable. As the semantic branch was trained on annotated data from Site~1, this evaluation is in-distribution: it confirms correct fusion behavior on the training site's failure modes and covers scenario types (e.g., Speedbreaker, Random LiDAR fail) not captured at Site~2. Table~\ref{table_fmea_sow} (unseen Site~2) is the primary evidence for cross-site generalization.}
\end{center}
\vspace{-1em}
\end{table}

\begin{table}[t]
\caption{FMEA Validation -- Site 2 (Unseen)}
\label{table_fmea_sow}
\vspace{-0.5em}
\begin{center}
\scriptsize
\begin{tabular}{|l|c|l|c|c|c|}
\hline
\textbf{Scenario} & \textbf{N} & \textbf{Camera} & \textbf{LiDAR} & \textbf{GT} & \textbf{Fused} \\
\hline
Wet mud & 4 & Wet Mud & T & NT & \textbf{NT} \\
Inclined slope & 11 & Flat Road & NT & T & \textbf{T} \\
Big puddle & 1 & Debris & T & NT & \textbf{NT} \\
Vegetation & 3 & Vegetation & T & NT & \textbf{NT} \\
Debris/rock & 9 & Debris & T & NT & \textbf{NT} \\
Stack of pipes & 9 & Pipes & T & NT & \textbf{NT} \\
Rebar stack & 2 & Rebar & T & NT & \textbf{NT} \\
Rocky road & 1 & Rocky Terrain & NT & T & \textbf{T} \\
Rocky terr.\ (misclass.) & 11 & Debris & T & T & NT (fail) \\
\hline
\end{tabular}
\end{center}
\vspace{-0.6em}
\begin{center}
\footnotesize{T = Traversable, NT = Non-Traversable.}
\end{center}
\vspace{-1em}
\end{table}

\section{LIMITATIONS AND FUTURE WORK}
\label{sec:limitations}

\textbf{Limitation:} On unseen Site~2, Rocky Terrain was frequently misclassified as Debris ($N{=}11$), likely due to overlapping visual texture between loose rock and construction debris, compounded by Rocky Terrain being underrepresented in Site~1-only training. LiDAR correctly read these cells as traversable, but Rule~5 is designed to trust high-confidence semantic evidence over LiDAR in exactly this geometry-clear, camera-flags-hazard scenario, so the high-confidence Debris prediction triggered the override and reported a false obstacle---the inverse of the intended correction, where the classification error propagates through the fusion policy rather than being suppressed by it. Debris also absorbs other similar hazards (e.g., \emph{Big puddle}), acting as a catch-all class; this points to a need for more diverse, cross-site Rocky Terrain (and broader hazard-class) annotations to improve semantic discrimination and reduce false semantic overrides, a use case for the released raw rosbag corpus. We also note the scope of this evaluation: as described in Section~\ref{sec:fmea}, the scenarios were selected because LiDAR alone fails on them, so they characterize failure-mode coverage rather than a general comparison against camera-only or uniform-override policies. Cell-level costmap metrics over full frames, sensitivity analysis on the gating thresholds, robustness under degraded vision (dust, low light, glare), and closed-loop navigation trials remain future work.

A further extension of the semantic taxonomy is the explicit representation of small sharp objects that may pose a puncture hazard to the AMR. Such objects can be difficult to identify reliably from LiDAR when their physical size is small relative to the sensor sampling density, so they produce insufficient geometric evidence and may appear as part of the surrounding ground surface, whereas their visual appearance may provide a stronger cue for camera-based semantic recognition. We therefore plan to introduce a dedicated \texttt{sharp\_object} class and include additional annotated examples in future dataset extensions. This would allow the semantic branch to identify localized puncture hazards that otherwise constitute geometric false negatives in the LiDAR-based traversability estimate, further extending the failure-mode-aware fusion strategy.

\section{CONCLUSION}

We presented a failure-mode-aware multimodal traversability system for construction AMRs that treats LiDAR and semantic vision as complementary rather than interchangeable: a deterministic, class- and confidence-gated fusion engine rejects visually hazardous but geometrically flat hazards and suppresses phantom obstacles on drivable elevation changes, while preserving flat road, terrain, and rocky terrain as distinct surface categories for downstream cost assignment, so that navigation can assign surface-specific traversal costs or speed preferences. We also release a multimodal construction-site dataset---four closed-loop ROS~2 sequences from two active sites plus a curated 506-frame, 28-class annotated subset---to support further construction perception, sensor fusion, traversability mapping, localization, loop-closure analysis, and navigation research. The complete system runs in real time on an NVIDIA Jetson AGX Orin, integrated with a custom AMR navigation stack: the standalone fine-tuned YOLOv26-m-seg model achieved a mean inference latency of 23.18~ms at $640\!\times\!640$ (43.14~FPS), and the integrated ROS~2 semantic--LiDAR pipeline a mean of 88.96~ms per update (11.24~Hz) against a 10~Hz costmap configuration, with P95 and P99 latencies above the 100~ms period.


\end{document}